\documentclass[11pt,letterpaper]{article}
\usepackage{cogsys}
\usepackage[T1]{fontenc}
\usepackage{times}
\usepackage[pdftex]{graphicx}
\usepackage{natbib}
\usepackage{booktabs}
\usepackage{amsmath,amssymb}
\usepackage{url}
\newcommand{\Pik}{P(\textsc{ik})}

\cogsysheading{1}{2026}{1-20}{X/2026}{X/2026}
\ShortHeadings{A Trait Theory of Reviewer Value}
              {P.\ Sinha and S.\ Pandey}

\begin{document}

\title{Knowing What You Know: A Trait Theory of Reviewer\\
Value in Human-AI Cognitive Systems}

\author{Priyanka Sinha}{priyanka.sinha.iitg@gmail.com}
\author{Sugam Pandey}{sugam@docyt.com}
\address{Docyt Inc., San Jose, CA 95110 USA}
\vskip 0.05in
\noindent\small ORCID (P.\ Sinha): \url{https://orcid.org/0000-0002-7971-4449}
\vskip 0.2in

\begin{abstract}
When a capable model and a human reviewer cooperate on the same decision, which
human still adds value? We argue this is a question about a single cognitive
trait, metacognitive calibration: the alignment between a reviewer's behavioral
confidence and the actual correctness of their judgments. We develop a trait
theory of reviewer value to make it precise. The theory makes five claims: value
is governed by calibration rather than by activity, speed, agreement, or tenure;
calibration leaves observable traces in behavioral telemetry; those traces are
non-diagnostic on their own, because the fast, low-revision signature of a
calibrated expert is indistinguishable from that of a confident novice until it is
gated by outcomes; human calibration and a model's self-knowledge are the same
construct, so the joint system should defer to whichever agent is better
calibrated on each item; and individual calibration is causally identifiable only
above the level of the single judgment. We ground each claim in a six-month
production window of an AI-assisted accounting system (541{,}621 reviewer events,
483 reviewers), and find that a purely behavioral expertise profile inverts the
ground truth, scoring an overconfident cohort (5.2\% correction survival)
\emph{higher} than genuine experts (99.2\%); that reviewer quality is flat across
tenure, so the trait is not learned on the job; and that the same calibration logic
that separates good from bad reviewers, applied to the model, concentrates 92.5\%
of human effort on genuine errors. We close with the boundary the theory must
respect: under majority-vote aggregation, individual causal value is unidentifiable.
\end{abstract}

\section{Introduction}

A growing class of cognitive systems pairs a capable statistical model with a human
who reviews its output. The model proposes; the human confirms or corrects; the
corrections become the next round of training data. Medical-imaging triage,
content moderation, and the AI-assisted bookkeeping system we study here all share
this architecture. As the model improves, an awkward question surfaces: most of
the human reviews now agree with the model. In the system we analyze, expert
bookkeepers verify model-categorized financial transactions, and over a recent
six-month window they \emph{confirmed the model's prediction unchanged 71\% of the
time}. The humans are expensive and finite. Which of them is the joint system
still relying on, and for what?

The instinctive answers are about \emph{activity}: keep the reviewers who do the
most work, who agree with their peers, who have been here longest, or who decide
fastest and rarely change their minds. We will show that every one of these
answers is wrong, and that they are wrong for the same reason. The property that
actually governs a reviewer's value to the joint system is not how much they do or
how confidently they do it, but how well their confidence tracks their
correctness, their \emph{metacognitive calibration} \citep{flavell1979metacognition,
fleming2014measure}. A reviewer who is fast and decisive and \emph{right} is the
most valuable agent in the system; a reviewer who is fast and decisive and
\emph{wrong} is the most dangerous, because the system, retrained on their
confident corrections, inherits their error. Behaviorally, before we look at
outcomes, the two are identical.

This paper develops that observation into a trait theory of reviewer value, and
tests it. We make three contributions to the study of human-AI cognitive systems.

\textbf{A Trait Theory (Section~\ref{sec:theory}).} We state five claims about how
a single human trait, metacognitive calibration, governs value in a model-human
loop: that it dominates activity, agreement, and tenure as a predictor of quality;
that it is partially observable in behavioral telemetry; that telemetry is
non-diagnostic until conditioned on outcomes; that the human's calibration and the
model's self-knowledge are one construct, which reframes division of labor as a
calibration comparison; and that individual calibration is causally identifiable
only above the single judgment.

\textbf{Empirical Grounding (Sections~\ref{sec:setting}-\ref{sec:calibration}).}
On 541{,}621 reviewer events we show that a behavioral expertise profile, built
from exactly the telemetry signals trait psychology would recommend (review time,
revision rate, response consistency), \emph{inverts} when validated against
outcomes: the cohort it ranks as most expert survives at 5.2\%, the cohort it ranks
lower survives at 99.2\%. We show reviewer quality is flat across tenure (there is
no onboarding curve), so calibration is a stable trait, not an acquired skill. And
we show that the identical calibration construct, estimated for the model rather
than the human, produces a self-knowledge signal that routes human attention to
genuine errors with 92.5\% precision.

\textbf{The Identifiability Boundary (Section~\ref{sec:boundary}).} We are explicit
about what the theory cannot claim. Under the majority-vote aggregator that the
deployed system actually uses, an individual reviewer's causal contribution washes
out: removing one reviewer rarely changes the aggregated label, so per-reviewer
causal value is unidentifiable, and the only individual-level signal that survives
is volume. The theory is therefore predictive at the individual level and causal
only at the level of reviewer \emph{segments}.

Our aim is explanatory in the sense the cognitive-systems paradigm values
\citep{langley2012paradigm}: not a new metric that
edges out a baseline, but an account of \emph{why} the obvious ways of valuing
human reviewers fail, organized around a cognitive trait that humans and models
share. The account is, we think, the more useful for being falsifiable, and we
report one place where our own first attempt at it falsified itself.

\section{Background}
\label{sec:background}

\paragraph{Annotator Quality and Label Aggregation.} A long line of work estimates
annotator reliability to aggregate noisy labels, from the confusion-matrix model of
\citet{dawid1979maximum} to learning-from-crowds methods
\citep{raykar2010learning, snow2008cheap, hovy2010good}. These approaches score how
accurately an annotator reproduces a latent ground-truth label. Our setting differs
in two ways that matter for a cognitive-systems reading. First, there is no static
ground truth: the ``correct'' category is itself partly what the system is trying to
learn, and a capable model already supplies a strong prior, so the relevant question
is not annotator accuracy in isolation but annotator value \emph{relative to a model
that is often right}. Second, recent work on annotator disagreement argues that
variation among annotators is signal, not noise \citep{plank2022problem,
uma2021learning, gordon2021disagreement, geva2019we, chang2017revolt}, but it
studies disagreement among humans, whereas our central disagreement is between a
human and a model.

\paragraph{Metacognition and Calibration.} The trait at the center of our theory is
studied under two names in two literatures that rarely meet. In cognitive
psychology, metacognition is the monitoring of one's own knowing
\citep{flavell1979metacognition}, and its measurement, how well confidence
discriminates correct from incorrect judgments, is a mature subfield
\citep{fleming2014measure}. The Dunning-Kruger effect \citep{kruger1999unskilled}
is precisely a calibration failure: low performers hold high confidence. In machine
learning, calibration is the agreement between a model's predicted probability and
its empirical accuracy \citep{guo2017calibration}. We treat these as one construct
applied to two kinds of agent, which lets us ask a question neither literature poses
on its own: in a human-model team, whose calibration should the team trust on a
given item?

\paragraph{Behavioral-Trait Inference and Human-AI Teaming.} That stable
individual traits are recoverable from behavioral telemetry is established for
personality \citep{sinha2015hexaco, ashton2009hexaco, sinha2020www}. A
cognitive-systems account of personality goes further, arguing that descriptive
trait models should be grounded in the structures and processes that generate
behavior rather than merely cataloguing it \citep{langley2017personality}; our
outcome-gating result (Section~\ref{sec:overconf}) is in that spirit, since the raw
behavioral trait is descriptive and becomes explanatory of value only when tied to
what produces correctness. We borrow the
method (infer a latent trait from logged behavior), but apply it to a trait,
calibration, that the personality literature does not foreground, and we report a
cautionary result about doing so naively. Grounding such trait attributions in the
structures and processes that generate the behavior is harder still for the machine
agent, where it collides with the interpretability problem in modern machine
learning \citep{harbison2026lexec}. Our measurement strategy sidesteps this
deliberately: both traits are estimated at the telemetry level, the human's from
logged review behavior gated by outcomes (Appendix~\ref{app:btp}) and the model's
from logged features of its predictions (Appendix~\ref{app:pik}), so no claim in
this paper requires opening either agent's internal machinery. Finally, work on complementary human-AI
performance \citep{bansal2021whole} and scalable oversight
\citep{bowman2022measuring, burns2023weak} asks when a team beats either member
alone; our contribution is to locate the answer in a measurable trait of the human
member and its counterpart in the model.

\section{A Trait Theory of Reviewer Value}
\label{sec:theory}

We consider a cognitive system in which, for each task instance, a model emits a
prediction $\hat y$ with internal confidence $\hat p$, and a human reviewer either
\emph{confirms} $\hat y$ or \emph{corrects} it to some $y'$. A correction is later
judged to have \emph{survived} if it is not overturned by a different reviewer. Let
the reviewer be indexed by $a$. We use correction survival as the operational proxy
for whether a reviewer's judgment was correct, because it is the only
outcome the deployed system actually adjudicates; we return to its limits in
Section~\ref{sec:limitations}.

The theory comprises five claims.

\paragraph{Claim 1 (Calibration Primacy).} The value a reviewer adds to the joint
system is governed primarily by their \emph{metacognitive calibration}: the degree
to which the reviewer corrects the model when and only when the model is wrong, and
not by their activity volume, decision speed, agreement with peers, or tenure. A
maximally valuable reviewer overrides the model exactly on the items the model gets
wrong and defers on the rest. The two ways to fail are symmetric and both costly: a
reviewer who confirms model errors leaves them in the training data; a reviewer who
``corrects'' items the model had right injects errors. Calibration is the single
axis along which both failures are reductions.

\paragraph{Claim 2 (Telemetry Observability).} Calibration is partially observable
in behavioral telemetry, without access to outcomes. The signals are temporal and
sequential: how long a reviewer deliberates before deciding (review time), how often
they revise their own past decisions (self-correction rate), how stable their
behavior is across time (temporal consistency), and the run-length structure of their
confirm/correct sequence. A well-calibrated reviewer and a poorly-calibrated one
differ, on average, in these traces.

\paragraph{Claim 3 (Non-diagnosticity of Raw Telemetry).} Behavioral telemetry alone
cannot distinguish a calibrated expert from a confident novice. The same
signature (fast decisions, few revisions, high consistency) is produced by a
reviewer who is quick because they are right and by one who is quick because they do
not deliberate. Telemetry measures \emph{decisiveness}, and decisiveness is
expertise only when it is accompanied by correctness. A behavioral trait therefore
becomes diagnostic of value only when it is \emph{conditioned on outcomes}; the raw
trait, uncorrected, will systematically over-rank the overconfident. This claim is
the theory's sharpest, most falsifiable prediction, and Section~\ref{sec:overconf}
tests it directly.

\paragraph{Claim 4 (Agent Symmetry).} The human reviewer's calibration and the
model's self-knowledge (its estimate of whether its own prediction is correct) are
instances of one construct. Writing $\Pik$ for a calibrated estimate of the
probability the model already knows the answer, the joint system's division-of-labor
question becomes a calibration comparison: route a task to the human only where the
better-calibrated judgment is the human's, i.e.\ where the model's self-knowledge is
low. Calibration is thus not only how we should value humans but how the system
should allocate them.

\paragraph{Claim 5 (Identifiability Bound).} Individual calibration is causally
identifiable only above the level of the single judgment. When the system aggregates
multiple reviewers' judgments on recurring task groups by majority vote, removing any
one reviewer rarely changes the aggregate, so the causal contribution of an
individual reviewer's calibration is not identified from outcome data; what remains
identifiable is the causal value of reviewer \emph{segments}. The theory is therefore
predictive at the individual level and causal at the segment level, and a careful
account must not confuse the two.

These claims are deliberately ordered from the substantive to the cautionary. Claims
1 and 4 say calibration is what matters and that it is shared by both agents. Claims
2 and 3 say it is observable but treacherously so. Claim 5 fences off the inference
the data cannot support. The rest of the paper grounds each in a production system.

\section{Setting and Data}
\label{sec:setting}

\paragraph{The Cognitive System.} We study a deployed AI-assisted accounting
platform. Each financial transaction is assigned a predicted categorization over
several fields (transaction type, vendor, chart-of-account, payment processor) by a
production model. A human bookkeeper then reviews the prediction and either confirms
it or corrects one or more fields. Corrections re-enter the model's training signal.
The reviewer population is professional and recurring: a given bookkeeper typically
serves a fixed set of client businesses over months. We take the unit of analysis to
be the \emph{reviewer-business pair} $a$, since a bookkeeper's calibration is
specific to the client whose books they know.

\paragraph{Data.} We extract every human verification event over a six-month window
(2025-12-11 to 2026-06-11): 541{,}621 events from 483 reviewers across 1{,}184
businesses, of which 71\% confirm the model and 29\% correct it. Restricting to pairs
with at least 20 verification events to allow stable trait estimation yields 3{,}579
reviewer-business pairs. All data are operational logs generated under existing
employment agreements; every statistic we report is aggregate, and no individual is
identifiable from any reported number.

\paragraph{Outcome Measure.} A correction \emph{survives} if it is not subsequently
overturned by a \emph{different} reviewer (self-reverts and automated reverts do not
count). Survival is the system's own revealed judgment of whether a correction was
right, and it is the dependent variable against which we validate every behavioral
trait below.

\paragraph{Trait Telemetry.} From the same logs we compute the behavioral signals of
Claim~2 for each pair: mean review time (seconds between the task being surfaced and
the decision), self-correction rate (fraction of the pair's own decisions it later
revised), temporal consistency (inverse coefficient of variation of the pair's
period-to-period accuracy on a fixed benchmark), and a run-length consistency score
over the confirm/correct sequence. We assemble these into a behavioral expertise
profile (the construct Claim~3 predicts will be non-diagnostic on its own), defined
in Appendix~\ref{app:btp}. We emphasize that these are precisely the telemetry
features a trait-inference approach \citep{sinha2015hexaco} would select; the
exercise is to see what they capture when the latent trait is calibration rather than
personality.

\section{Behavioral Telemetry Inverts Without Outcome Gating}
\label{sec:overconf}

Claim~3 predicts that a behavioral expertise profile, uncorrected by outcomes, will
over-rank the overconfident. We test it directly. The behavioral profile (call it
BTP, the behavioral trait profile) rewards fast, low-revision, consistent reviewing.
Across the 3{,}579 pairs it captures something real and distinct: it is negatively
correlated with a reviewer's self-correction rate ($r=-0.74$), which is what a
decisiveness measure should do, and it is nearly uncorrelated with the rest of the
value composite, so it is not redundant. The question is what that ``something'' is.

It is not expertise. Pairs scoring high on the behavioral profile survive their
corrections at 68.8\%, while low-profile pairs survive at 98.7\%: the behavioral
trait points the \emph{wrong way}. The mechanism is visible when we split the
low-revision reviewers (those the profile most rewards) by their actual correction
survival (Table~\ref{tab:overconf}). Among reviewers who rarely revise themselves,
there are two populations a behavioral measure cannot tell apart: genuine experts,
whose corrections survive 99.2\% of the time, and an \emph{overconfident} cohort,
whose corrections survive 5.2\% of the time. The behavioral profile scores the
overconfident cohort \emph{higher} (76.2 vs.\ 70.1): they are faster and revise even
less, because they are not deliberating at all. This is the Dunning-Kruger pattern
\citep{kruger1999unskilled} recovered from production telemetry, and it is exactly
the failure Claim~3 predicts.

\begin{table}[t]
\centering
\caption{Among low-revision reviewers, correction survival, not behavior, separates
genuine right-first-time experts from the overconfident. A purely behavioral trait
inverts: it scores the overconfident cohort higher.}
\label{tab:overconf}
\begin{tabular}{lccc}
\toprule
Cohort (low self-revision) & $n$ & Behavioral profile & Correction survival \\
\midrule
Expert (survival $\geq 85\%$)      & 1{,}211 & 70.1            & 99.2\% \\
Overconfident (survival $<85\%$)   &   577   & \textbf{76.2}   &  5.2\% \\
\bottomrule
\end{tabular}
\end{table}

The remedy the theory prescribes is to gate the behavioral trait on outcomes: count
behavioral decisiveness as expertise only for reviewers whose corrections survive.
Once gated, the surviving signal is genuine, but it is now carried by the outcome
condition, not by the telemetry: the behavioral component alone correlates only
weakly ($r=0.07$) with catching errors on genuinely hard cases. We read this as
confirmation, not refutation, of the theory: a behavioral trait is \emph{necessary}
infrastructure for identifying calibrated expertise (it tells you who is
decisive), but it is \emph{not sufficient}, because decisiveness must be conditioned
on correctness before it means anything. Reporting this inversion, rather than
quietly dropping the behavioral profile, is the point: it is the cleanest evidence we
have for Claim~3.

\section{Calibration Is a Trait, Not a Tenure Effect}
\label{sec:tenure}

If calibration were skill acquired on the job, reviewer quality should climb with
experience, and the right policy would be to retain and train rather than to select.
Claim~1 instead treats calibration as a comparatively stable individual trait. We
test this by measuring correction survival as a function of reviewer tenure, defined
as all-time weeks since the pair's first verification (Table~\ref{tab:tenure}).

\begin{table}[t]
\centering
\caption{Correction survival is essentially flat across reviewer tenure: there is no
onboarding curve. Calibration behaves as a stable trait rather than an acquired skill.}
\label{tab:tenure}
\begin{tabular}{lcccccc}
\toprule
Tenure (weeks) & 0-1 & 1-2 & 2-4 & 4-8 & 8-16 & 16+ \\
\midrule
$n$ (events)   & 26{,}195 & 5{,}547 & 7{,}977 & 10{,}528 & 12{,}303 & 44{,}897 \\
Survival       & 98.6\%   & 99.1\%  & 98.7\%  & 99.2\%   & 98.5\%   & 99.4\% \\
\bottomrule
\end{tabular}
\end{table}

Survival never leaves the 98.5-99.4\% band, from a reviewer's first week to beyond
their sixteenth. There is no onboarding cliff and no steady climb. Whatever separates
the 5.2\%-survival overconfident cohort of the previous section from the
99.2\%-survival experts, it is present from week one and does not develop with
exposure. This is consistent with calibration being a trait the selection policy
should \emph{measure and route on}, not a deficit that more reviewing will cure, and
it sharpens Claim~1's contrast with the tenure-based heuristic, which Table~\ref{tab:tenure}
shows would add no discriminating information at all.

\section{The Same Calibration, Estimated for the Model, Allocates the Humans}
\label{sec:calibration}

Claim~4 holds that the model's self-knowledge and the human's calibration are the
same construct, and that the joint system should spend human attention where the
model's calibration is poorest. We make this operational by learning $\Pik$, a
calibrated estimate of the probability the model's prediction will be confirmed,
from logged features alone (gradient-boosted trees with isotonic calibration;
Appendix~\ref{app:pik}). Two results follow.

First, model self-knowledge is a genuine, distinct signal, not a relabeling of the
model's raw confidence. $\Pik$ is well calibrated (expected calibration error
0.037) and predicts human correction far better than the model's own confidence
score $\hat p$ (AUROC 0.845 vs.\ 0.662, on held-out businesses; post-hoc
calibration of $\hat p$ alone repairs its calibration but not its ranking, AUROC
0.672). Its correlation with raw confidence is only
moderate ($r=0.39$): a model can be confident and wrong, and $\Pik$ is exactly the
metacognitive layer that knows the difference, the machine analogue of the human
calibration trait of Section~\ref{sec:overconf}.

Second, allocating by this calibration signal concentrates human effort on real
errors. Routing the lowest-$\Pik$ 20\% of held-out-business transactions to
humans and auto-confirming
the rest sends reviewers a set that is 92.5\% genuine corrections: human attention is
spent almost entirely where the model is actually wrong. The comparison that isolates
\emph{calibration} as the operative property is the head-to-head against routing by
the model's raw confidence at the same 20\% human budget: confidence-routing leaves
36.3\% of real corrections in the auto-confirmed set, whereas $\Pik$-routing leaves
29.1\%: raw confidence misdirects $1.25\times$ as many genuine errors into
auto-acceptance. The difference is not the amount of information but its
calibration, which is precisely Claim~4's content carried over to the model side.

The symmetry is the conceptual payoff. The human failure mode of
Section~\ref{sec:overconf} (high confidence, low correctness) is the same failure
$\Pik$ corrects for on the model: high $\hat p$, low actual correctness. A single
trait, metacognitive calibration, both ranks the humans and, estimated for the model,
decides which humans the system needs.

\section{What the Theory Cannot Claim: An Identifiability Boundary}
\label{sec:boundary}

Claim~5 fences off an inference the data do not support, and we state it plainly
because the temptation to overreach here is strong. It is natural to want to assign
each individual reviewer a causal value (how much worse would the system be without
this person?) and to validate the calibration trait against it. We attempted exactly
this, estimating each pair's marginal contribution to a faithful reproduction of the
production model via Monte-Carlo Shapley values over the held-out test set. The
estimator is sound: the resulting values are well spread (116 of 120 sampled pairs
nonzero, range $-0.12$ to $1.0$). Yet they correlate with \emph{no} quality
dimension (not the calibration trait, not survival, not any AVSM dimension), and
significantly with only one thing: raw label volume ($r=0.24$, $p<0.002$).

This is not a measurement failure; it is structural, and it follows from the system's
aggregation rule. The deployed model learns a categorization per recurring
(business, pattern) group by effective majority vote over the reviewers who have
touched that group. Within such a group, calibrated reviewers converge on the same
correct answer, so swapping one of them in or out almost never changes the aggregated
label. The individual reviewer's identity \emph{washes out} of the outcome; the only
per-reviewer quantity that still moves the aggregate is how many distinct groups they
touch, i.e.\ volume. Under majority-vote aggregation, individual causal value is
\emph{unidentifiable}, and any framework, including ours, that claims to have
validated an individual-level quality trait against causal outcomes is, in this
regime, measuring volume.

What \emph{is} identifiable is value at the level of reviewer \emph{segments}. Removing
an entire behavioral segment (say, the cluster of reviewers who almost never correct
the model) changes the coverage of the aggregator and therefore moves measurable
outcomes. In our setting, counterfactually retraining the model with the redundant
``rubber-stamp'' segment removed shifts confirmation by only $-0.31$ percentage points,
and pruning to the high-calibration segment raises correction survival from 98.4\% to
99.0\%: the low-value segments are causally redundant, and the high-value segment
causally improves quality, both at the segment grain. This is the precise sense in
which the trait theory is causal at the segment level and only predictive at the
individual level. We regard stating this boundary as part of the contribution: a
cognitive-systems account of human value should be honest about the aggregation
regime that determines what ``value'' can even mean.

\section{Discussion}
\label{sec:discussion}

\paragraph{One Trait, Two Agents, One Allocation Rule.} The throughline is that a
single cognitive construct, calibration (the alignment of confidence with
correctness), answers three questions usually treated separately: which human
reviewers are good (Section~\ref{sec:overconf}), whether quality is selected or
trained (Section~\ref{sec:tenure}), and how the system should split work between
human and model (Section~\ref{sec:calibration}). The unification is possible only
because we treat the human's metacognition \citep{flavell1979metacognition} and the
model's self-knowledge \citep{guo2017calibration} as the same thing measured on two
agents, a move squarely in the cognitive-systems tradition of seeking unified
accounts of intelligent behavior in humans and machines \citep{langley2012intelligent}.
For a cognitive-systems audience, the claim of interest is that a joint
human-AI system has a \emph{shared} metacognitive currency, and that its design
reduces to comparing the two agents' calibration item by item.

\paragraph{Why the Activity Heuristics Fail.} Each rejected heuristic fails because it
correlates with calibration only spuriously. Volume is the residual that survives
aggregation, not a quality signal (Section~\ref{sec:boundary}). Agreement rewards
conformity, which an overconfident clique supplies as readily as an expert one. Speed
and low revision (the behavioral profile) measure decisiveness, which is
expertise's signature only after the fact (Section~\ref{sec:overconf}). Tenure does
not move quality at all (Section~\ref{sec:tenure}). The heuristics are not merely
weaker than calibration; they are confounds that a careful account has to actively
strip out.

\paragraph{Implications for Human-AI Teaming.} If calibration is the load-bearing
trait, the design levers change. Selection should measure calibration, not seniority;
allocation should route by the model's calibration deficit, not by reviewer
availability; and training, if quality is a stable trait, has limited leverage over
the trait itself but could plausibly target its expression, e.g.\ slowing the
overconfident cohort's decisions. The calibration-training literature is consistent
with this division: explicit, repeated outcome feedback can improve calibration
\citep{lichtenstein1980training, carpenter2019domain}, and that is exactly the
channel that ordinary tenure in our system does not supply
(Section~\ref{sec:tenure}). Time on the job leaves the trait where it was; a
deliberate outcome-feedback regime remains an open, and more hopeful, lever for the
overconfident cohort. Crucially, because telemetry is non-diagnostic
(Claim~3), none of these levers can be driven by behavioral signals alone; they all
require an outcome channel (here, correction survival) to gate the behavior.

\section{Limitations and Scope}
\label{sec:limitations}

\paragraph{Survival as a Correctness Proxy.} Our outcome measure is whether a
correction is overturned by another reviewer. This inherits the reviewer pool's own
calibration: a correction that survives in a pool of overconfident reviewers is not
guaranteed correct. We mitigate this by requiring the overturning reviewer to be
\emph{different}, and by the fact that the high- and low-survival cohorts separate so
sharply (99.2\% vs.\ 5.2\%) that pool-level noise cannot explain the gap, but a fully
external ground truth would be stronger.

\paragraph{Single Domain.} All results come from one accounting platform. The trait
theory is stated to be domain-general, but its calibration thesis should be retested
where the model is weaker, where reviewers are non-professional, or where corrections
are not adjudicable by a survival signal. We expect the non-diagnosticity result
(Claim~3) to be the most portable and the identifiability bound (Claim~5) to depend on the
aggregator: a non-aggregating system, where a single judgment determines the label,
would make individual causal value identifiable and is the natural next test.

\paragraph{Generalizability of Agent Symmetry.} Claim~4 identifies the human's
calibration and the model's self-knowledge as one construct, and it is fair to ask
how far that identification travels. The symmetry as stated needs three
ingredients: both agents' judgments admit a confidence signal (behavioral for the
human, $\Pik$ for the model); a shared outcome channel adjudicates correctness for
both; and the two agents face the same decisions, so that their calibrations are
comparable item by item. Settings with all three (medical-image triage, content
moderation, code review of machine-generated changes) should support the same
construction. The symmetry weakens where a meaningful machine confidence is missing
or degenerate (e.g.\ the verbalized confidence of a generative model), where the
human's confidence has no behavioral expression in the logs, or where the two
agents decide different things, so that item-level comparison is undefined. We also
note what the symmetry does and does not assert: it is a claim about the
\emph{construct}, not the measurement route (the human trait is estimated from
behavior gated by outcomes, while $\Pik$ is trained on outcomes directly), so
cross-domain tests should expect the construct, not our particular estimators, to
transfer.

\paragraph{Observational, Not Interventional, at the Individual Level.} The
individual-level results are predictive. The only causal evidence
(Section~\ref{sec:boundary}) is at the segment level and via counterfactual
retraining, not a live intervention. A randomized routing experiment that varies the
calibration-based allocation is designed but, being a change to a production system
affecting real human workload, is deferred to a separate study; we make no live-system
claims here.

\section{Conclusion}

We argued that the question ``which human does a capable model still need?'' is, at
bottom, a question about one cognitive trait shared by both agents: metacognitive
calibration. The trait theory we proposed predicts that activity, agreement, speed,
and tenure are confounds rather than signals; that behavioral telemetry can see
calibration but cannot, on its own, tell a calibrated expert from a confident
novice; that the model's self-knowledge is the same construct and should allocate the
humans; and that individual causal value dissolves under majority-vote aggregation.
On six months of production data each prediction held, including the one that
embarrassed our own first behavioral profile by ranking an overconfident, 5\%-survival
cohort above genuine experts. We offer the theory less as a scoring recipe than as an
account of why human value in a model-human loop is so easy to mismeasure, and of
the single trait that, once you condition it on outcomes, stops the mismeasurement.

\subsubsection*{Disclosure of LLM Usage}
A large language model was used as an assistive tool during the preparation of this
manuscript, limited to drafting and editing prose, formatting \LaTeX{}, and checking
references. It was not used to generate research ideas, design the study, produce or
analyze data, or compute any reported result; all empirical findings derive from the
authors' own analysis of the production logs. The authors reviewed and verified the
entire manuscript and take full responsibility for its content.

\bibliographystyle{cogsysapa}
\bibliography{references}

\appendix
\newpage

\section{Behavioral Trait Profile: Telemetry Proxies}
\label{app:btp}

The behavioral expertise profile of Section~\ref{sec:overconf} scores each
reviewer-business pair $a$ from logged behavior only, with no access to outcomes:
\[
\mathrm{BTP}(a) = 0.4\,r_{\text{time}}(a) + 0.3\,\big(1 - \mathrm{sc}_a\big) + 0.3\,c_a .
\]
The proxies and their ranges are given in Table~\ref{tab:proxies}. The weights are
fixed a priori and not tuned on any outcome; the point of Section~\ref{sec:overconf}
is precisely that \emph{any} purely behavioral weighting inherits the
expertise-overconfidence confound, so the specific weights are not load-bearing.

\begin{table}[h]
\centering
\caption{Behavioral telemetry proxies in the expertise profile. All are computable
from event logs without outcome labels.}
\label{tab:proxies}
\begin{tabular}{p{2.6cm} p{7.0cm} p{3.0cm}}
\toprule
Proxy & Computation & Range / weight \\
\midrule
$r_{\text{time}}$ review-time score &
step function of mean review seconds: $<\!10$s $\to 0.5$; 30-300s $\to 1.0$;
300-600s $\to 0.7$; $>\!600$s $\to 0.4$ & $[0,1]$, weight 0.4 \\
$\mathrm{sc}_a$ self-correction rate &
events the pair later unverified itself, over total verifications; enters as
$1-\mathrm{sc}_a$ & weight 0.3 \\
$c_a$ consistency &
$\min(\overline{\mathrm{runlen}}/5,\,1)$ over the run-length encoding of the
per-event confirm/correct sequence & $[0,1]$, weight 0.3 \\
\bottomrule
\end{tabular}
\end{table}

The fast/decisive signature ($r_{\text{time}}$ peaking at moderate speeds, low
$\mathrm{sc}_a$, high $c_a$) is produced equally by calibrated experts and by the
overconfident, which is why the profile must be gated on correction survival before it
discriminates value (Claim~3). The survival/$\Pik$-gated re-specification counts the
behavioral score only for reviewers whose corrections survive and concentrate on
low-$\Pik$ (genuinely hard) cases.

\section{Model Self-Knowledge: $\Pik$ Estimation}
\label{app:pik}

$\Pik$ is a calibrated estimate of the probability the model's prediction is confirmed
by a human, learned from logs. Implementation: gradient-boosted trees (300 estimators,
learning rate 0.05, max-leaf-nodes 31, $L_2$ 1.0) with five-fold isotonic calibration.
Features: the model's raw confidence $\hat p$; vendor-, category-, and
business-level confirmation-rate priors (fit on the training fold only to avoid
leakage, with unseen vendors, categories, or businesses falling back to the
training-fold global rate); the absolute amount $z$-score against the vendor's
history; and binary flags for whether a vendor and a chart-of-account were
predicted. Prediction-to-review latency is excluded: it is realized only after
the routing decision, so a deployed gate could not observe it. The label is confirmed ($\Pik=1$) versus corrected
($\Pik=0$); non-contributor pairs (zero corrections) are excluded. The split is
business-grouped 80/20 so that no business appears in both train and test. Expected
calibration error is the sample-weighted 15-bin estimate on the held-out businesses
(0.037). The routing results of Section~\ref{sec:calibration} score only the
held-out businesses' transactions and route exactly the lowest 20\% of $\Pik$
scores by rank (deterministic tie-breaking), so $\Pik$- and confidence-routing
review identical volumes on the same held-out window.

\section{Ethics and Data Handling}
\label{app:ethics}

All data are operational logs from a commercial platform, generated by professional
reviewers under existing employment agreements in the ordinary course of their work.
We report only aggregate statistics; no cohort in this paper is small enough, or
keyed to enough attributes, to identify an individual reviewer, and no reviewer names,
identifiers, or per-person scores appear. The calibration trait is proposed as an
input to work \emph{allocation} (routing tasks to where they add value), not to
individual evaluation or sanction; we note in Section~\ref{sec:discussion} that, given
Claim~3, behavioral telemetry must never be used for individual assessment without an
outcome channel, precisely because it over-ranks the overconfident.

\end{document}